\documentclass[
sigconf = true, 
review = false, 
screen = true, 
anonymous = false, 
nonacm = true
]{acmart}

\usepackage{amsmath,amsthm}
\usepackage{mathtools}
\usepackage{wasysym} 
\usepackage{fontawesome}
\usepackage{tikz}
\usepackage{tkz-euclide}
\usepackage{xspace}
\usepackage{epsfig}
\usepackage{color}
\usepackage{url}
\usepackage{graphicx,dsfont}
\usepackage{tikz}
\usepackage{float}
\usepackage{natbib}
\usepackage{longtable}
\usepackage{booktabs}
\usepackage{makecell}
\usepackage{multicol}
\usepackage{multirow}
\usepackage{colortbl}
\usepackage{tikz}
\usepackage{csquotes}
\usepackage{algpseudocode}
\usepackage[vlined,linesnumbered,ruled,titlenotnumbered]
{algorithm2e}

\usepackage{marginnote}

\usepackage{pifont}

\usepackage{color}

\newcommand\ignore[1]{}

\begin{document}

\title[Generating Instances with Performance Differences for More Than Just Two Algorithms]{Generating Instances with Performance Differences\\for More Than Just Two Algorithms}


\author{Jakob Bossek}
\affiliation{%
  \institution{Dept. of Information Systems}
  \city{University of M\"unster, Germany}
  \country{}
}
\email{jakob.bossek@wi.uni-muenster.de}

\author{Markus Wagner}
\affiliation{%
  \institution{School of Computer Science}
  \city{The University of Adelaide, Australia}
  \country{}
}
\email{markus.wagner@adelaide.edu.au}

\begin{abstract}
In recent years, Evolutionary Algorithms (EAs) have frequently been adopted to evolve instances for optimization problems that pose difficulties for one algorithm while being rather easy for a competitor and vice versa.
Typically, this is achieved by either minimizing or maximizing the performance difference or ratio which serves as the fitness function.
Repeating this process is useful to gain insights into strengths/weaknesses of certain algorithms or to build a set of instances with strong performance differences as a foundation for automatic per-instance algorithm selection or configuration.

We contribute to this branch of research by proposing fitness-functions to evolve instances that show large performance differences for more than just two algorithms simultaneously.
As a proof-of-principle, we evolve instances of the multi-component Traveling Thief Problem~(TTP) for three incomplete TTP-solvers.
Our results point out that our strategies are promising, but unsurprisingly their success strongly relies on the algorithms' performance complementarity.

\end{abstract}

\maketitle


\section{Introduction}
\label{sec:introduction}


The usefulness of benchmarking suites is affected by four characteristics~\cite{bartzbeielstein2020benchmarking}: the instance set should be diverse, representative, scalable and tunable, and knowing the best solutions (or at least the best performance) is beneficial as well. With our research, we target the ``diversity'' criterion, as this is a characteristic that is not always immediately obvious from a quick inspection, in contrast to, for example, scalability.

Instances can be diverse in the feature space and in the performance space. In the former, the focus is on covering a range of different problem characteristics; in the latter, the focus is on generating instances on which algorithms behave differently.
Instances with different characteristics can not only support our studies of problem difficulty and lead to new approaches to the problem, but they can also act as a tool for training and using per-instance algorithm configurators~\cite{Hutter2007aac} or algorithm selectors~\cite{Kerschke2019aacSurvey}.

To create instances with desired characteristics, the manual generation is often labour-intensive, if not practically impossible: it requires a substantial amount of domain knowledge together with in-depth knowledge of the target solver and advanced mathematical skills, especially when dealing with randomized heuristics.
Moreover, as such instance engineers are (anecdotally) scarce, the approach with the human-in-the-loop is often impractical.
Hence, to explore the space of instances in an automatic way, evolutionary algorithms are often used. Among the first to do so was \citet{Hemert2003Evolving}, who initially evolved difficult instances for the binary constraint satisfaction problem and later for other combinatorial problems as well.
Since then, many other researchers have explored a number of research directions that build upon this general idea of instance evolution:
(1) transfer of the concept to other problem domains,
(2) creation of instances to broadly cover the feature space, and
(3) evolving instances on which pairs of solvers behave as differently as possible (i.e., instances would be difficult for one solver but easy for the other).

With this present article, we address an open problem in the third research direction: how can we evolve instances for sets of more than just two algorithms?
This extension is not trivial, as one needs to consider performance rankings for $N$ solvers as well as the discriminatory performance aspect.
To achieve this, we introduce several fitness functions that can guide evolutionary algorithms in the generation of instances for more than two algorithms simultaneously.
As a proof-of-principle, we generate instances for the permutation-based Travelling Thief Problem, assess the effectiveness of our approaches, and raise awareness for a number of issues that researchers might come across when conducting similar studies in the future.





\section{Related work}
\label{sec:related_work}



In recent years Evolutionary Algorithms (EAs) have frequently been adopted to evolve instances for optimization problems.
Among the first was \citet{Hemert2003Evolving} who evolved difficult instances for the binary constraint satisfaction problem.
Later, this work was extended~\cite{10.1162/evco.2006.14.4.433} to the generation of hard instances for binary constraint satisfaction, Boolean satisfiability and the traveling salesperson problem (TSP), where he also pointed out structural properties that make instances hard for certain algorithms.

Interestingly, the TSP has been a popular domain for instance generation, possibly because instances could easily inspected by humans, and because the definition of instance features based on point clouds (i.e. the locations of cities) is relatively easy in itself.
What has changed over time for the TSP were the fitness functions
-- first, the number of local search operations as a proxy for difficulty~\cite{SmithMilesH2011Discovering},
then approximation quality~\cite{MersmannBTWBN13TSPfeaturebased,MBB2012LocalSearchTSP}, and
also multiple objectives~\cite{10.1007/978-3-319-13563-2_19} --
and the heuristics -- Bossek~et~al.~\cite{DBLP:conf/lion/BossekT16, BossekT2016UnterstandingTSP} started to target state-of-the-art heuristics with small instances, and employed disruptive operators in order to be able to evolve instances for current state-of-the-art heuristics with strong feature diversity without relying on explicit diversity preservation~\citet{DBLP:conf/foga/BossekKN00T19}.

Other domains besides the TSP that have also been subject to the evolution of instances; they include the knapsack problem~\cite{Plata2019knapsackEvol}, the quadratic knapsack problem~\cite{Julstrom2009Evolving}, the graph colouring problem~\cite{bowly2013evolving}, and the Hamiltonian completion problem~\cite{lechien2021evolving}.

Most of these mentioned works had as the primary goal the evolution of instances on which a pair of solvers behaves differently.
As this purely performance-focused view at diversity has the potential of resulting in sets of highly similar instances, research has begun to incorporate the diversity of features into the process.
Again, the TSP has been the first target~\cite{DBLP:conf/ppsn/GaoNN16}, where diverse instance sets (with respect to a single selected feature at a time) could be achieved in the evolution.
Along similar lines, but for machine learning problems, it was shown that it is possible to guide the evolution of an instance to a particular target vector in a high-dimensional space~\cite{Munoz2018vector}. A very similar approach was recently demonstrated for continuous black-box optimization problems~\cite{Munoz2020coBBO}.
To address the challenge of simultaneously achieving diverse sets with respect to multiple features, several schemes were proposed that operate in the multi-dimensional feature space and are thus independent of the problem domain~\cite{Neumann2018discrepancy,Neumann2019diversity}.

In summary, while the field has come a long way, it still seems to be an open problem of how one can efficiently generate instance sets for more than two solvers.



\section{Problem formulation}
\label{sec:problem_formulation}

The major scheme of this paper is the process of evolving problem instances for a combinatorial optimization problem. We are given a set of $N$ algorithms $\mathcal{A}=\{A_1,\ldots,A_N\}$. We denote by $p_A(I)$ the performance of algorithm $A$ on input instance $I$ (w.l.o.g. we assume the performance to be maximized). If a subset of the algorithms is randomized we assume that $p$ is some adequate aggregated performance such as the median performance score. In order to keep formulas clean we often identify algorithms by natural numbers, i.e. $\mathcal{A}=\{1, \ldots, N\}$ and we write short $p_i$ if the instance can be derived easily from the context.
Note that for $N$ algorithms there are $N!$ possible \emph{rankings} of the algorithm performances $p_1, \ldots, p_N$, e.g. for $N=3$ we have $p_1 \geq p_2 \geq p_3$, $p_1 \geq p_3 \geq p_2$, $\ldots$, $p_3 \geq p_2 \geq p_1$. The goal is to generate a set of $L$ instances where (approximately) $L/N!$ of the instances follow each of the $N!$ rankings and in addition, the performance of the algorithms on each instance differ substantially. Hence, we aim for instances with \emph{diverse} algorithmic rankings.

\section{Evolving instances}
\label{sec:evolving_instances}

We approach the setting formulated in Section~\ref{sec:problem_formulation} by adopting a simple $(1+1)$-EA which is outlined in Algorithm~\ref{alg:evolving_ea}. The EA first generates a random problem instance $I$. Next, a copy of $I$ is subject to mutation. The resulting instance $I'$ is compared to $I$ by means of a suitable fitness function $F$. If $I'$ is no worse than $I$ with respect to fitness, it replaces $I$, otherwise it is discarded. This simple random process is repeated until some stopping condition, usually a generous time limit, is met. Certainly, initialization, variation and many implementation details are highly problem-dependent. In addition, the success depends on the fitness function that serves as a driver for the process and ideally guides the EA towards better instances.
\begin{algorithm}[t]
\SetKwInOut{Input}{input}
\Input{Fitness function $F$}
Initialize instance $I$ randomly\;
\While(\Comment{Often time-limit}){budget not depleted}{
    Generate $I'$ by applying mutation to $I$\;
    \If{$F(I') \geq F(I)$}{
        Replace $I$ with $I'$\;
    } 
} 
\Return{$I$}\;
\caption{Outline of instance evolving $(1+1)$-EA.}
\label{alg:evolving_ea}
\end{algorithm}


\section{Fitness functions to guide the EA}
\label{sec:fitness_functions}

In existing work the goal was to either (a) generate instances that are particularly difficult to solve for a single solver $A$ (e.g., \cite{Hemert2003Evolving, 10.1162/evco.2006.14.4.433}) or (b) easy for one solver $A$ and hard(er) for a competitor $B$ (e.g., \cite{MersmannBTWBN13TSPfeaturebased}, \cite{DBLP:conf/foga/BossekKN00T19}) by means of evolutionary search. Note that for both goals, the fitness function that guides the EA is rather straight-forward and natural. For option (a) we may take the ratio $p_A(I) /OPT$ where $OPT$ is an optimal solution to the problem and maximize this ratio. For option (b) we may maximize $p_A(I)/p_B(I)$ to guide the EA towards instances which are easier for $A$ and maximize the reciprocal to obtain instances that are harder for $A$. Likewise we can maximize the difference $p_A(I) - p_B(I)$ the one or the other way around. Recall that for $N=2$ there are just two possibilities (neglecting the case of equal performance): either $A$ is better than $B$ or $B$ is better than $A$. For the general case with $N \geq 3$ there are $N!$ possibilities we aim to cover.\footnote{While having a single outstanding solver per instance is a good goal for the purpose of eventual algorithm selection, full permutations can enable deeper analytical dives to understand when algorithms perform mediocrely or badly.} Thus, there is no single straight-forward way to directly transfer the notion of \enquote{direction} to the general case. In the following we discuss three ways to generate a balanced set of $L$ instances with performance differences for $N \geq 3$ algorithms.

\paragraph{Pairwise approach} This naive approach relies on option (b) for two algorithms discussed above. Consider all $N(N-1)$ ordered pairs of algorithms $(i, j)$ and evolve instances that are easy for $A_i$ and hard for $A_j$ by maximizing the respective performance difference or ratio. The obvious drawback of this approach is that we have no influence on the performance of all other algorithms in $\mathcal{A} \setminus \{i,j\}$ on the evolved instance which may hinder the aimed diversity of the instance set. In fact, the evolved problem can be very easy (or hard) for all those algorithms. This is undesirable.

\begin{figure}[t]
\centering
\begin{tikzpicture}[scale=1]
\draw (-1.5,0) edge[dotted] (6.5,0);
\draw (-1,0) edge[solid] (6,0);
\foreach \x in {0.3,3.4,4.9}
\draw (\x,0.1) edge[solid] (\x,-0.1);
\node (l) at (0.3,-0.5) {$p_{(1)}$};
\node (c) at (3.4,-0.5) {$p_{(2)}$};
\node (r) at (4.9,-0.5) {$p_{(3)}$};

\draw [decorate,decoration={brace,amplitude=3pt},yshift=5pt] (0.3,0) -- (3.4,0) node [above, midway, yshift=5pt] {$(p_{(2)} - p_{(1)})$};
\draw [decorate,decoration={brace,amplitude=3pt},yshift=5pt] (3.4,0) -- (4.9,0) node [above, midway, yshift=5pt] {$(p_{(3)} - p_{(2)})$};
\draw [decorate,decoration={brace,amplitude=3pt},yshift=5pt] (0.9,0.85) -- (5.1,0.85) node [above, midway, yshift=5pt] {$F(p_1, p_2, p_3) = (p_{(2)} - p_{(1)}) \cdot (p_{(3)} - p_{(2)})$};
\end{tikzpicture}
\caption{Illustration of no-order fitness calculation.}
\label{fig:fitness_no_order}
\end{figure}
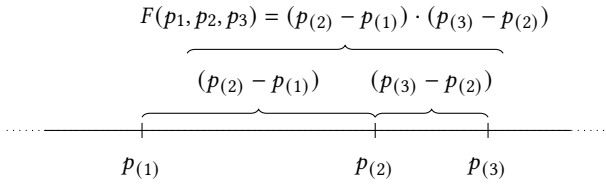

\paragraph{No order} This approach ignores explicit rankings. Instead, the performance values $p_1, \ldots, p_N$ are sorted in increasing order such that $p_{(1)} \leq p_{(2)} \leq \ldots \leq p_{(N)}$ where $p_{(i)}$ is the $i$-\emph{th} order statistic. Eventually, the fitness is calculated as
\begin{align}
\label{eq:fitness_no_order}
F(p_1, \ldots, p_N) = \sum_{i=2}^{N-1} (p_{(i)} - p_{(i-1)}) \cdot (p_{(i+1)} - p_{(i)}).
\end{align}
See Figure~\ref{fig:fitness_no_order} for an example.
A similar approach was used by Gao~et~al.~\cite{DBLP:conf/ppsn/GaoNN16} to evolve TSP instances with diverse feature values in the context of evolutionary diversity optimization.

This approach is simple, but in order to obtain a set with each $L/N!$ instances of an explicit ranking we need to be lucky since again there is no way to explicitly enforce a certain ranking.

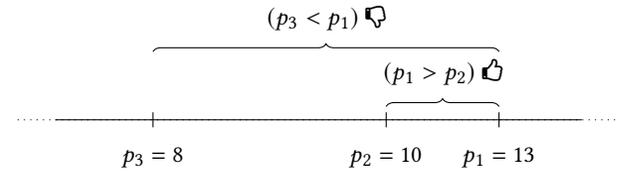
\begin{figure}[t]
\centering
\begin{tikzpicture}[scale=1]
\draw (-1.5,0) edge[dotted] (6.5,0);
\draw (-1,0) edge[solid] (6,0);
\foreach \x in {0.3,3.4,4.9}
\draw (\x,0.1) edge[solid] (\x,-0.1);
\node (l) at (0.3,-0.5) {$p_3=8$};
\node (c) at (3.4,-0.5) {$p_2=10$};
\node (r) at (4.9,-0.5) {$p_1=13$};

\draw [decorate,decoration={brace,amplitude=3pt},yshift=5pt] (3.4,0) -- (4.9,0) node [above, midway, yshift=5pt] {$(p_1 > p_2)$ \faThumbsOUp};
\draw [decorate,decoration={brace,amplitude=3pt},yshift=5pt] (0.3,0.73) -- (4.9,0.73) node [above, midway, yshift=5pt] {$(p_3 < p_1)$ \faThumbsODown};
\end{tikzpicture}
\caption{Illustration of explicit ranking fitness calculation with $\pi=(3,1,2)$.}
\label{fig:fitness_explicit_ranking_a}
\end{figure}

\paragraph{Explicit ranking} The name already suggests the idea. Here, the EA explicitly tries to establish a certain parameterizable ranking / performance permutation $\pi = (\pi(1), \ldots, \pi(N))$ in a two-phase approach. The first phase aims to come up with an instance with the desired ranking, while the second phase aims to maximize the performance difference once the ranking is achieved. Formally, let $p_1, \ldots, p_N$ be the performance values of the algorithms and let $\pi$ describe the desired ranking. Let $G = \{(i, i+1) \, | \, p_{\pi(i)} \geq p_{\pi(i+1)}\}$ and $B = \{(i, i+1) \, | \, p_{\pi(i)} < p_{\pi(i+1)}\}$ be the set of \emph{\textbf{g}ood} and \emph{\textbf{b}ad directions} respectively; i.e. $G$ contains all pairs of algorithms which are in the right order according to $\pi$ whereas the pairs in $B$ violate $\pi$. With this notation the goal is to maximize the vector-valued fitness function
\begin{align}
\label{eq:fitness_explicit_ranking}
F(p_1, \ldots, p_n; \pi) = \left(|G|, f_B, f_G\right)
\end{align}
in \emph{lexicographic order} where
\begin{align*}
f_B =
\begin{cases}
\sum_{(i,j) \in B} (p_{\pi(i)} - p_{\pi(j)}) & \text{ if } |B| > 0 \\
0 & \text{ otherwise},
\end{cases}
\end{align*}
and
\begin{align*}
f_G =
\begin{cases}
\sum_{(i,j) \in G} (p_{\pi(i)} - p_{\pi(j)}) & \text{ if } |G| > 0 \\
-\infty & \text{ otherwise}.
\end{cases}
\end{align*}
The first component is simply the number of good directions $|G| \in \{0,1,\ldots,N-1\}$. The second component is given by the sum of differences between pairs of bad directions. Hence, by definition of $B$, every single additive term is negative and so is the sum in case $B$ is non-empty. Once there are no bad directions anymore, the value takes its maximum value zero. Note that once this happens $|G|=N-1$ and $f_B=0$ hold and the EA will not accept any solution which is lexicographically inferior in subsequent iterations (the first two components will not change anymore). The last component, $f_G$, adds up the performance differences of good directions and becomes relevant once the desired ranking is reached. For example, consider $p_1 = 13, p_2 = 10, p_3 = 8$ and $\pi = (3, 1, 2)$. Then $G = \{(2, 3)\}$ because $p_{\pi(2)} = p_1 \geq p_2 = p_{\pi(3)}$ and likewise $B = \{(1,2)\}$ (see Figure~\ref{fig:fitness_explicit_ranking_a}). The fitness vector is hence $F(p_1, p_2, p_3; \pi) = (1, -5, 3)$. Now consider another instance with performance values $p'_1 = 13, p'_2 = 10, p'_3 = 15$. It follows $B = \emptyset$ and in consequence $F(p'_1, p'_2, p'_3; \pi) = (2, 0, 5)$. Thus, the instance resulting in the latter performance values would be accepted.

We stress that for this approach to be successful it must be possible to achieve the desired ranking. This requires a portfolio of algorithms where there is no strongly dominating algorithm which outperforms its competitors by orders of magnitude on (almost) all instances. If such an algorithm existed, it would be impossible to achieve a ranking in phase one where it would perform worst. We will get back to possible pitfalls and issues we experienced during (preliminary) experiments later in Section~\ref{sec:results}.


\section{Proof-of-concept study}
\label{sec:proof_of_concept_study}

In the upcoming two sections -- as a proof of concept -- we adopt our approach to generate instances with performance differences for three heuristics ($N=3$) for the Traveling Thief Problem.

\subsection{The Traveling Thief Problem}

Real-world optimization problems often consist of several $\mathcal{NP}$-hard combinatorial optimization problems that interact with each other~\citep{klamroth2016interwoven,Bonyadi2019}. Such multi-component optimization problems are difficult to solve not only because of the contained hard optimization problems, but in particular, because of the interdependencies between the different components. Interdependence complicates decision-making by forcing each sub-problem to influence the quality and feasibility of solutions of the other sub-problems. Examples of multi-component problems are vehicle routing problems under loading constraints~\citep{iori2010routing,pollaris2015vehicle}, maximizing material utilization while respecting a production schedule~\citep{cheng2016supply,wang2020integrated}, and relocation of containers in a port while minimizing idle times of ships~\citep{forster2012tree,jin2015solving,hottung2020deep}.

In 2013, \citet{bonyadi2013travelling} introduced the Traveling Thief Problem (TTP) as an academic multi-component problem. The academic `twist' of it is particularly important because it combines the classical Traveling Salesperson Problem (TSP) and the Knapsack Problem (KP) -- both of which are very well studied in isolation -- and because of the interaction of both components can be adjusted. 

\paragraph{Formal Definition.}
We are given a set of
$n$ cities, the associated matrix of distances $d_{ij}$, and
a set of $m$ items distributed among these cities. Each item $k$ is defined by
a profit $p_{k}$ and a weight $w_k$. A thief must visit all the cities exactly once, stealing
some items on the road, and return to the starting city.

The knapsack has a capacity limit of $W$, i.e. the total weight of the collected items must not exceed $W$.
In addition, we consider a renting rate $R$ that the thief must pay at
the end of the travel, and the maximum and minimum velocities denoted
$v_{max}$ and $v_{min}$ respectively. Furthermore, each item is available in only one city,
and $A_i \in \{1, \dots, n\}$ denotes the availability vector. $A_i$ contains
the reference to the city that contains the item $i$.

A TTP solution is typically coded in two parts: the tour $X = (x_1, \dots, x_n)$,
a vector containing the ordered list of cities, and the picking plan
$Z = (z_1, \dots, z_m)$, a binary vector representing the states of items
($1$ for packed, $0$ for unpacked).

To establish a dependency between the sub-problems, the TTP was designed such that the speed of
the thief changes according to the knapsack weight. To achieve this, the thief's velocity at
city $c$ is defined as
$v_x = v_{max} - C \times w_x$,
where $C = \frac{v_{max}-v_{min}}{W}$ is a constant value, and $w_x$ is the weight of the knapsack
at city~$x$.

The total value of items is
$g(Z) = \sum_m p_m \times z_m$, such that $\sum_m w_m \times z_m \le W $.
The total travel time is
$f(X, Z) = \sum_{i=1}^{n-1} t_{x_i, x_{i+1}} + t_{x_n, x_1}$,
where $t_{x_i, x_{i+1}} = \frac{d_{x_i, x_{i+1}}}{v_{x_i}}$ is the travel time from $x_i$ to $x_{i+1}$.

The TTP's objective is to maximize the total travel gain function,
which is the total profit of all items minus the travel time multiplied with the renting rate:
$F(X, Z) = g(Z) -  f(X, Z) \times R$.

For a worked example, we refer the interested reader to the initial TTP article by Bonyadi et al.~\cite{bonyadi2013travelling}.

\paragraph{TTP Solvers.}
The TTP has been gaining attention due to its challenging interconnected multi-component structure, and also propelled by several competitions 
organized to solve it, which have led to significant progress in improving the performance of solvers. Among these are iterative and local search heuristics~\citep{polyakovskiy2014comprehensive,maity2020efficient}\ignore{faulkner2015approximate},
solution approaches based on co-evolutionary strategies~\citep{bonyadi2014socially,el2015cosolver2b,namazi2019cooperative}, memetic algorithms~\citep{mei2014interdependence,el2016population}, swarm-intelligence approaches~\citep{Zouari2019antstpp,wagner2016stealing}, simulated annealing~\citep{el2018efficiently} and estimation of distribution approaches~\citep{Martins2017ttpeda}. Exact approaches were considered, however, they are limited to address very small instances~\citep{wu2017exact,neumann2017ttpPTAS}\ignore{wu2018evolutionary}. Moreover, dynamic TTP variants have been explored~\cite{sachdeva2020dynamic,herring2020dynamic}, as well as various multi-objective formulations~\cite{blank2017solvingBittp,yafrani2017ttpemoOLD,wu2018evolutionary,chagas2020nondominated}.
To better understand the effect of operators 
on a more fundamental level, fitness-landscape analyses~\cite{elyafrani2018ttplandscape,Wuijts2019ttpinvest} presented correlations and characteristics that are potentially exploitable.

\paragraph{TTP Instances.}
Almost all articles known to the authors rely on the $9\,720$ instances introduced by~\citet{polyakovskiy2014comprehensive} in 2014\footnote{\url{http://cs.adelaide.edu.au/~optlog/CEC2014COMP_InstancesNew/}} -- a small number of other instances is either created randomly or by following the scheme in~\cite{polyakovskiy2014comprehensive}.
Even though \citet{polyakovskiy2014comprehensive} created them systematically and with the intention to ``keep a balance between two components of the problem'', an inspection of the good solutions created across various papers reveals that they appear to greatly favour near-optimal TSP tours over near-optimal KP packing plans. This in turn seems to often affect the design decisions that an algorithm's creator makes; for example, many of the above-mentioned approaches create a good TSP tour first -- and independent of the KP/TTP -- as a starting point, and only then consider both interdependent components together; the other way around, i.e. starting with a good packing plan and then trying to make it work with a tour has not yet been fruitful, to the best of our knowledge.

In our opinion, this bias limits algorithm development as well as research on inter-dependencies, which the TTP is supposed to facilitate in the first place. Instance generation for the TTP -- which has not been done before, and which we use here for a proof-of-principle -- can thus open up opportunities for future research, as we will then be able to create instance sets specialized for the investigation of performance differences of single (or multiple) algorithmic design decisions.

\subsection{Experimental setup}

\paragraph{Heuristics $\mathcal{A}$} We select the following three heuristics from \cite{faulkner2015approximate} due to their high similarity as well as due to their structural differences:
\begin{itemize}
\item S2: run Chained Lin-Kernigham~(CLK), then PackIterative, then repeat Bitflip until converged;
\item S4: run CLK, then PackIterative, then repeat Insertion until converged;
\item C2: run CLK, then PackIterative, then repeat ``one Bitflip pass, one (1+1)-EA pass, one Insertion pass''.
\end{itemize}

PackIterative is a fast, mostly constructive packing heuristic
that takes into account the items' values and weights as well as the distance that they have to travel to the end along the given tour. Bitflip and (1+1)-EA operate exclusively on the packing plan, and either toggle the packing status in a deterministic fashion, or toggle the status of each item with probability $1/m$. Similarly, Insertion searches deterministically over the TSP part of a TTP solution by enumerating permutation-based insertions.

The rationale is as follows. Even though all three algorithms share the same first phase, the subsequent iterative hill-climbing differs in an important aspect: the simple heuristic S2 focuses exclusively on the packing plan, S4 focuses exclusively on the tour, and the more complex C2 incorporates components of both S2 and S4. 
While C2 can be seen as generally superior to the other two, there is the potential for C2 to be outperformed depending on the structure of the instance that strictly ``favours'' one problem component over the other. However, as it is apriori unclear what such instances would have to look like in order to put C2 at a disadvantage (when compared to the simpler S2 and S4), we leave it up to the evolution to tackle this challenge.

\begin{figure*}[ht]
\centering
\includegraphics[width=0.49\linewidth]{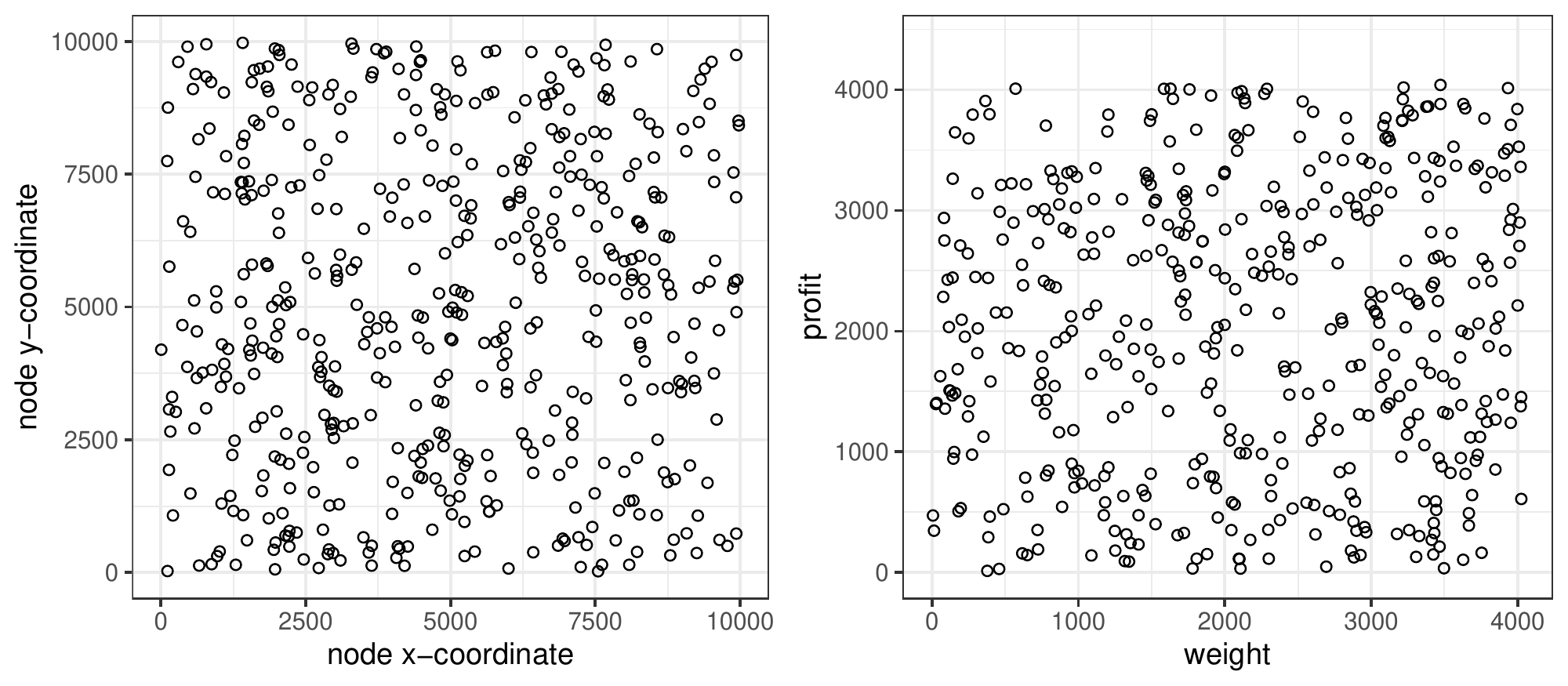}
\hfill
\includegraphics[width=0.49\linewidth]{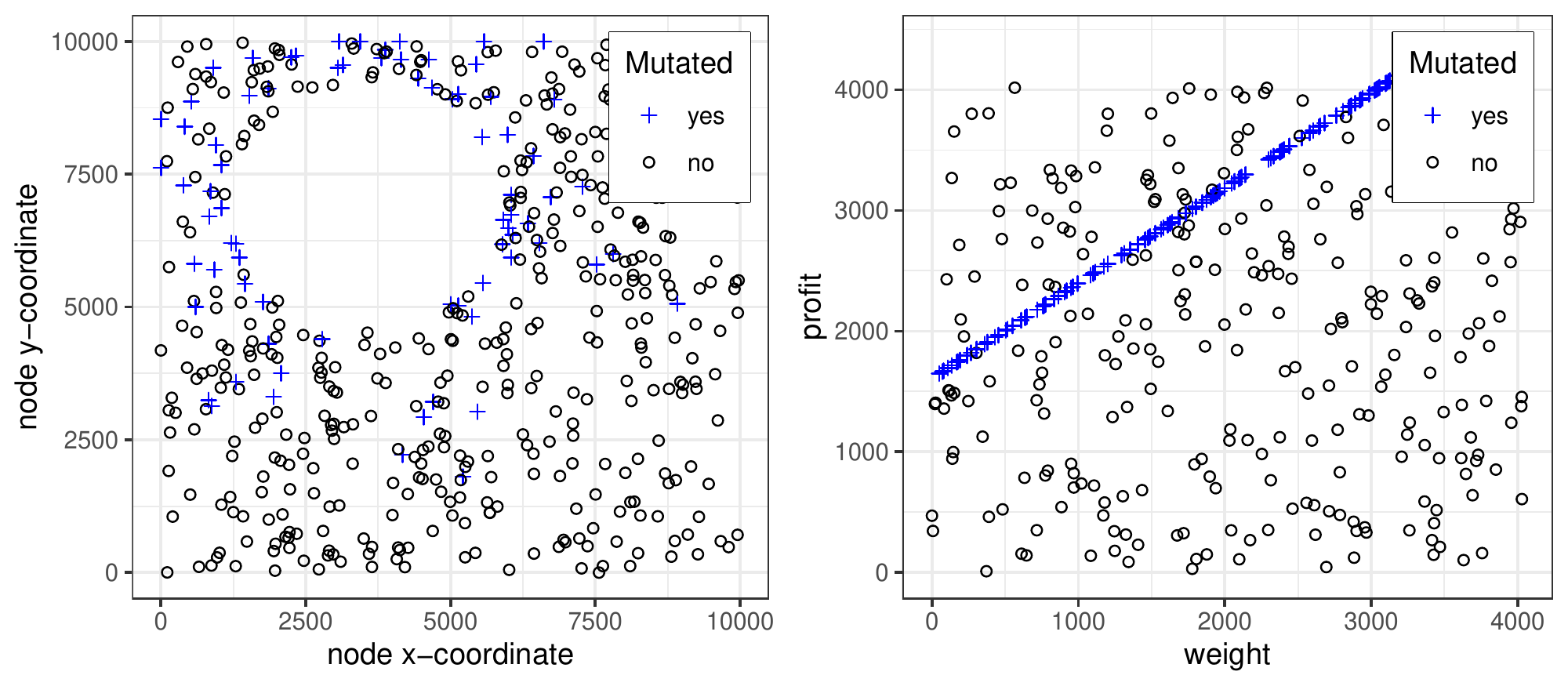}\vspace{-3mm}
\caption{An initial random TTP instance (left plots) and mutation applied to it (explosion mutation to the node coordinates, linear projection mutation to the weight-profit combination; right plots).}\vspace{-1mm}
\label{fig:EA_mutation_example}
\end{figure*}

\paragraph{EA components} Due to the large number of components of the TTP problem, the EA operators are quite involved. The EA is initialized with a random TTP instance with $n$ nodes and $\mathit{IPN}$ number of items per node. Random in this context means that $n$ points are placed uniformly at random in the Euclidean sub-plane $[0, 10\,000]^2$ to account for the TSP-component.
Moreover, the renting rate $R$ is chosen uniformly at random from the real-valued interval $[0, 1\,000]$ to allow for the influence of the travel time on the overall objective score to vary from small to large; this interval's upper bound is the result of considering the maximum value of $R$ across already known TTP instances and then increasing it further to allow for a broader range of interdependence.
Item weights are sampled from $[0, 4\,040]$ and profits are sampled from $[0, 4\,400]$.\footnote{These ``unusual'' ranges are an artefact of the knapsack generator by \citet{Martello99} that \citet{polyakovskiy2014comprehensive} used: said generator multiplies the generated weights and profits by up to a factor of four (in order to create harder ``cores'' of the knapsacks), which can be observed in the $9\,720$ TTP instances, however, this detail had not been reported by \citet{polyakovskiy2014comprehensive}, who only mention the intended upper bounds of $\sim1\,000$, and not the actual $\sim4\,000$.}
Lastly, and in line with \cite{polyakovskiy2014comprehensive}, the initial knapsack capacity is $W = \lceil(D/11) \cdot \sum_{m} w_m\rceil$ with $D$ chosen uniformly at random from $[1, \ldots, 10]$, and the minimum and maximum speeds are kept constant at $v_{min}=0.1$ and $v_{max}=1$.

We build upon work in the context of instance generation for the TSP and adopt mutation operators introduced in~\cite{DBLP:conf/foga/BossekKN00T19}. The authors proposed a set of mutation operators that aim for rather extreme changes to the node coordinates. This was motivated by the fact that earlier work in the context of evolutionary-guided TSP instance generation, using just small local changes to node coordinates, used to produce instances that in fact showed the aimed strong performance difference, but did not differ much from random uniform instances in terms of visual structure and instance characteristics. One illustrative example for the \enquote{creative} mutation operators is the explosion mutation where a center of explosion $c$ and an explosion radius $r>0$ are sampled and all points within Euclidean distance at most $r$ from $c$ are moved away from the explosion center. For an exhaustive description of all operators we refer the interested reader to~\cite{DBLP:conf/foga/BossekKN00T19}. Note that weights and profits $(w_i,p_i), 1 \leq i \leq m$ can be interpreted as a point cloud, too. Therefore, we apply the same operators to these values. Figure~\ref{fig:EA_mutation_example} illustrates an exemplary initial TTP instance and a mutant based on two different mutation operators. Here, the rather disruptive nature of the mutation operators becomes obvious.
We apply Gaussian mutation to the renting rate $R$, i.e., we add a random number stemming from a normal distribution with mean value zero and standard deviation~10 to it. Mutation of the knapsack capacity $W$ follows its random initialization scheme discussed above and is thus more disruptive. All variation operations are finalized with a repair step (random re-positioning within the bounds) that ensures that the respective coordinates/points/parameters stay within their initially defined bounds~(see initialization). 
In each iteration, the EA chooses one out of 10 mutation operators with equal probability to generate a mutant where all components are subject to mutation in every iteration.

\paragraph{Further parameters} We consider $n=200$ nodes and $\mathit{IPN} \in \{1, 3, 5, 10\}$ items per node. With respect to the proposed fitness functions we generate for each combination of $n$, $\mathit{IPN}$ and fitness function approach $240$ instances: for the two fitness functions where rankings actually matter, the set contains each 10 instances for each of the 24 $(n, \mathit{IPN}, \text{ranking})$-combinations. In total the generated benchmark set contains $N=720$ instances.
Within the fitness function evaluation each relevant heuristic is run $k=5$ times independently with a time-limit of $10$ seconds per run.\footnote{Actually, this time-limit is never hit because solvers terminated after at most one second.} The performance of an algorithm is defined as the unique median over all $k$ runs.\footnote{Actually, we would have liked to increase $k$ to 10 or even 30. However, in order to obtain a reasonable number of iterations of the evolving EA within the wall-time of each job, we relied on this rather small value.}
Each run of the evolving EA is given a wall-time of 48 hours as the single termination criterion: the EA terminates after $\approx 47$ hours and the last hour is used to run each of the three TTP heuristics 30 times independently on the evolved TTP instance for final evaluation. All experiments were run on the High-Performance-Cluster <anonymous>. We implemented the EA in the statistical programming language R~\cite{Rlang} in version 4.0.0; for the TTP heuristics we rely on existing Java implementations kindly provided by the original authors of~\cite{faulkner2015approximate}. All scripts and data are made available in a public GitHub.\footnote{Code and data: \url{https://github.com/jakobbossek/GECCO2021-ECPERM-ttp-evolving}}


\section{Discussion of results}
\label{sec:results}

We now discuss observations based on detailed data analysis of the generated instance set. Saying it right away: the results in the considered TTP setting are less pleasing than expected and may thus partially be considered as negative results. However, we have plausible explanations for these artifacts and feel like the lessons learned in the course of preparing this manuscript are of high scientific interest for researchers in the field.

\subsection{Desired versus actual rankings}

\begin{figure}[t]
\centering
\includegraphics[width=0.95\columnwidth]{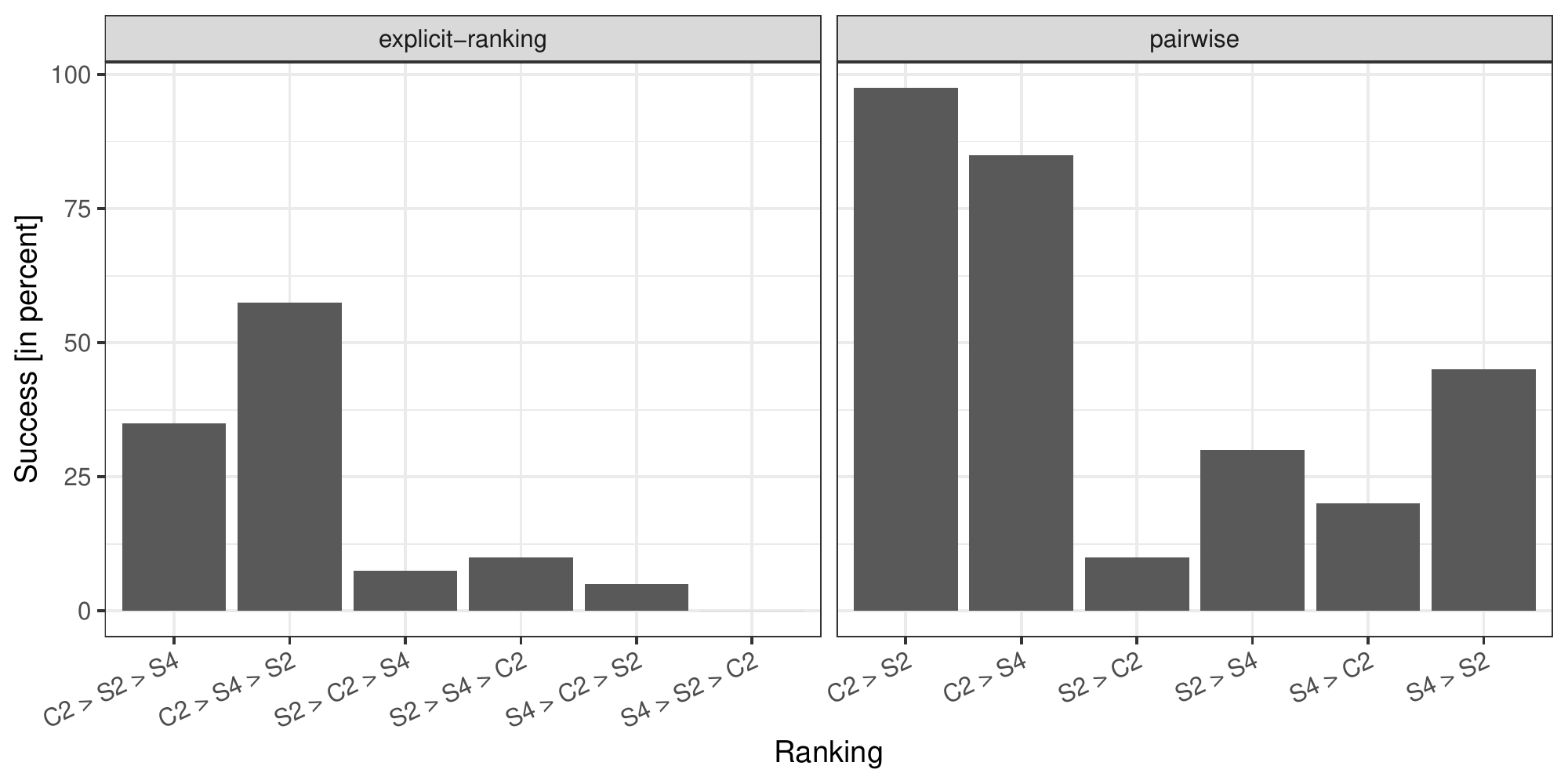}\vspace{-3mm}
\caption{Percentage of successful evolving jobs, i.e., jobs where the median of the final 30~evaluations in fact reflects the ranking that was aimed for.}\vspace{-4mm}
\label{fig:barplot_ranking_achieved}
\end{figure}

We first consider the results for the fitness function \emph{pairwise} and \emph{explicit-ranking} where the ranking is a parameter of the fitness function. Figure~\ref{fig:barplot_ranking_achieved} shows the rate of \enquote{successful} jobs for each ranking for the pairwise and explicit-ranking setting. Successful in this context means that the final evaluations reflect the desired ranking used by the fitness function.
We observe a clear trend. In the pairwise-setting (right plot in Figure~\ref{fig:barplot_ranking_achieved}) the EA succeeds in $85\%$ and even $97.5\%$ of the cases in generating instances where C2 performs better than S4 and S2 respectively. All other cases are far less successful. In particular, it seems difficult to evolve instances where S2 or S4 outperform C2 (success rates of $10\%$ and $20\%$ respectively). This observation indicates that C2 is clearly dominating the algorithm portfolio and can be attributed to the more sophisticated working principles of C2 in direct comparison to S2 and S4. Looking at the left plot in Figure~\ref{fig:barplot_ranking_achieved} we see that this issue directly transfers to the success rate of jobs guided by the explicit-ranking fitness function. This is plausible: if S2 $>$ C2 is hardly possible, we cannot expect S2 $>$ C2 $>$ S4 to be any easier.

\begin{figure}[t]
\centering
\includegraphics[width=0.95\columnwidth]{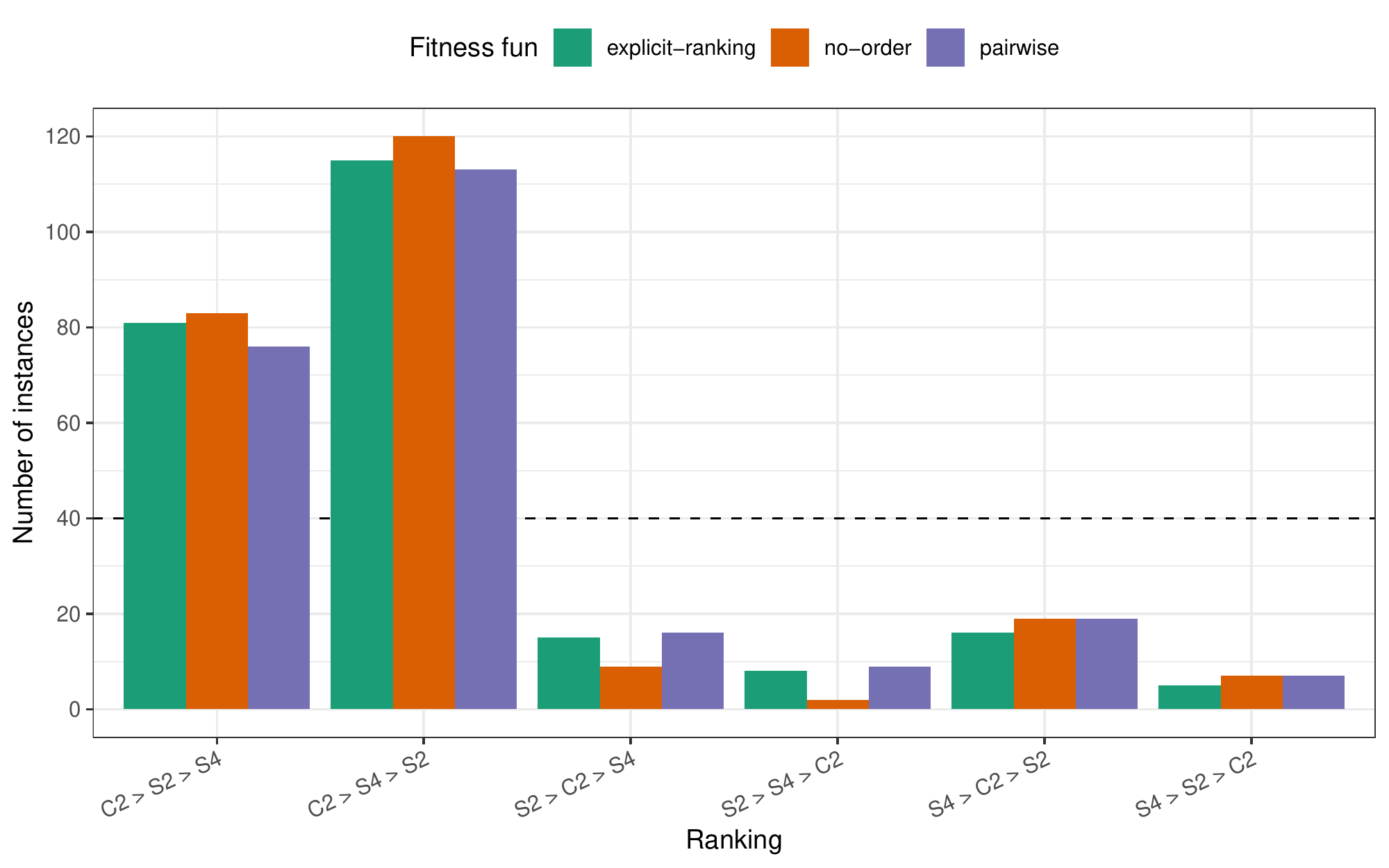}\vspace{-3mm}
\caption{Number of instances evolved for each ranking split by fitness-function. The dashed line indicates the number of instances that the experimental setup aimed for for each ranking.}\vspace{-2mm}
\label{fig:barplot_actual_ranking}
\end{figure}

Figure~\ref{fig:barplot_actual_ranking} shows a different perspective. Here, the ranking on the $x$-axis is the \emph{actual ranking} based on the final evaluations after the evolving process completed. This allows to integrate the results for the third fitness function, no-order, into the plot. The plot reveals that most evolved instances are easiest for C2 with each more than 110 instances in total showing the ranking C2 $>$ S4 $>$ S2 and about 80 instances reflecting C2 $>$ S2 $>$ S4. Instances where algorithms S2/S4 score first place are rather scarce (far less than the anticipated 40~instances for each ranking). Note however, that at least for S2 the results for the pairwise and explicit-ranking are superior to the plain no-order setting.

\subsection{Investigating issues}

\begin{figure*}[ht]
\centering
\includegraphics[width=\linewidth, trim=0 16pt 0 0, clip]{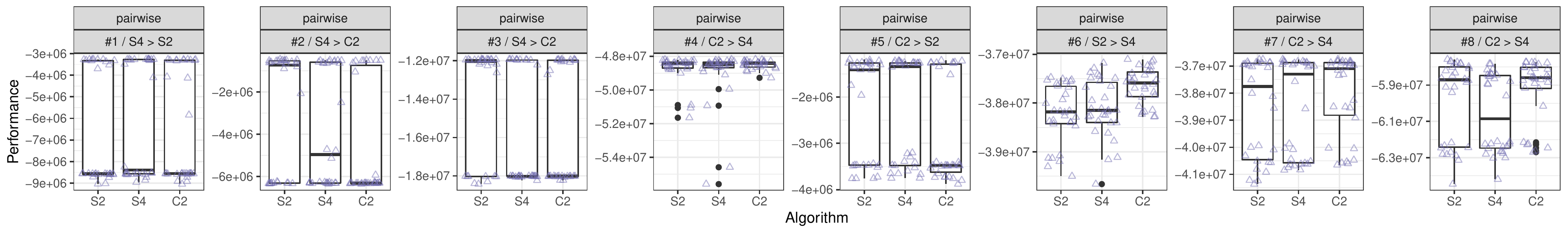}
\includegraphics[width=\linewidth, trim=0 16pt 0 0, clip]{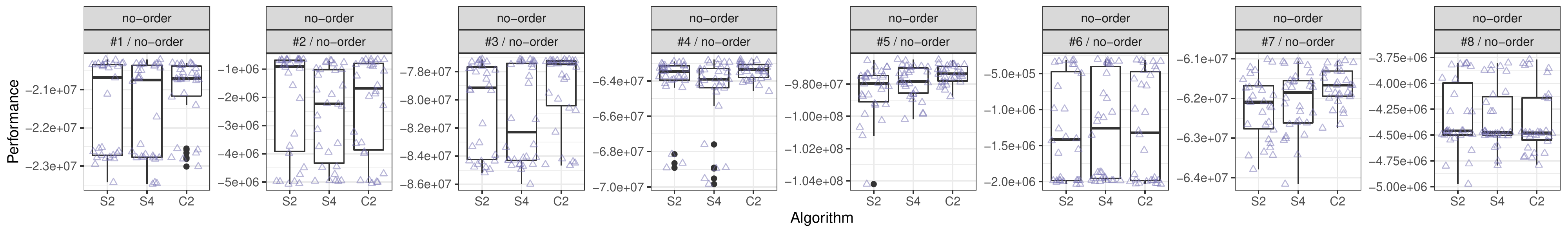}
\includegraphics[width=\linewidth]{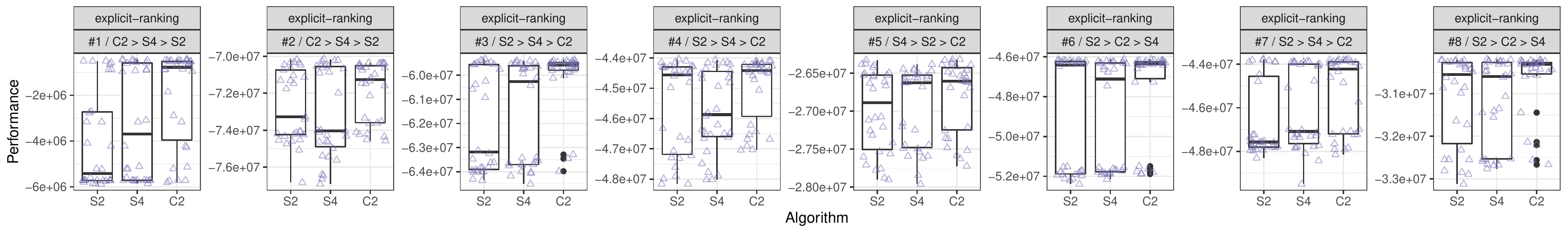}
\vskip-10pt
\caption{Distribution of performance values for each eight instances evolved by fitness function pairwise (top row), no-order (middle row) and explicit-ranking (bottom row).}
\label{fig:boxplots_evals}
\end{figure*}

We now take a closer look at the performance of the algorithms on a representative subset of the evolved instances. Figure~\ref{fig:boxplots_evals} shows boxplots of the performance of all three algorithms (each 30 runs) on eight representative evolved instances for each of the three fitness functions: pairwise, no-order and explicit ranking from top to bottom. We can identify two repeating patterns that pose difficulties to the EA; each alone can fool the EA and in particular the combination of both aspects. The first one is due to (roughly) bi-modal behavior of all or a subset of the algorithms. We can observe this, e.g. in the first three plots of the first row, the first plot in the second row and the sixth plot in the third row of (blue triangles indicate the raw performance values). Here, the algorithms seemingly run in either one of two local optima or at least multiple optima with very similar objective scores with roughly equal probability. Recall that the EA works with aggregated median performance values (based on each $k=5$ independent runs) and the fitness value in all cases is a composition of differences of median values. Now, for example assume, that algorithm S2 takes the raw performance values $(10, 10, 1, 1, 10)$ on some instance $I$: the median value is 10. For S4 we have $(10, 1, 1, 10, 1)$ and the median value is~1. In the pairwise setting for the ranking S2 $>$ S4 the fitness value would be $10-1=9$ and the instance might get accepted as the new incumbent instance in the course of optimization. Let us assume that this incumbent is not replaced anymore in subsequent iterations of the EA and the EA returns $I$. Now, due to the described issue, in the final 30 evaluations, the median difference on $I$ might be exactly the other way around indicating S4 $>$ S2. This is what actually happens very often. The other aspect is a high variance of objective scores intermingled with the fact that there are few cases where one algorithm always outperforms its competitors. In fact, on almost all instances, the best objective scores of the worst algorithm (with respect to median performance) are equal to the best scores of the best algorithm. In combination, both discussed issues have the potential to misguide the EA. In fact, looking a the development of the fitness values over time reveals that all EA-runs make progress \enquote{in the right direction}. However, final evaluations in most cases show the vice-versa. One could argue that these issues might by overcome by increasing the number $k$ of runs of each algorithm in the fitness-function evaluation. However, follow-up investigations with $k=30$ did not change the overall picture.

\subsection{Properties of evolved instances}

Next, we characterise the so-called features of the evolved instances. Useful features describe high-level properties of an instance that are (a) computationally undemanding to calculate (at least in comparison to costly runs of optimization algorithms) and (b) well-suited to distinguish algorithm performance. These features can then ideally be used in the context of automatic per-instance algorithm selection to predict the most likely best-performing algorithm from a portfolio (see~\cite{KerschkeHNT19AlgSelSurvey} for a recent survey on algorithm selection).

\begin{figure*}[ht]
\centering
\includegraphics[width=0.32\linewidth]{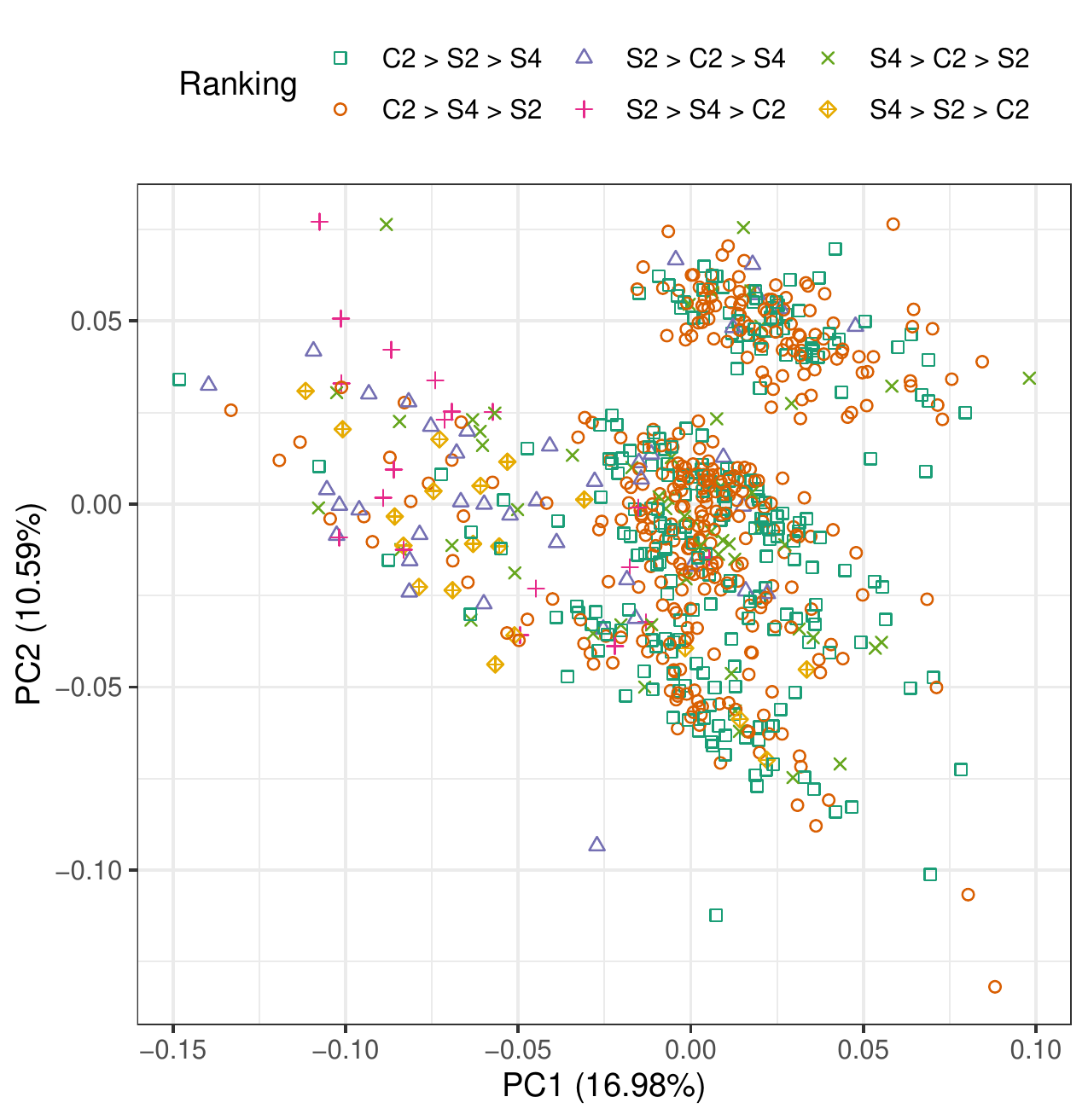}
\hfill
\includegraphics[width=0.32\linewidth]{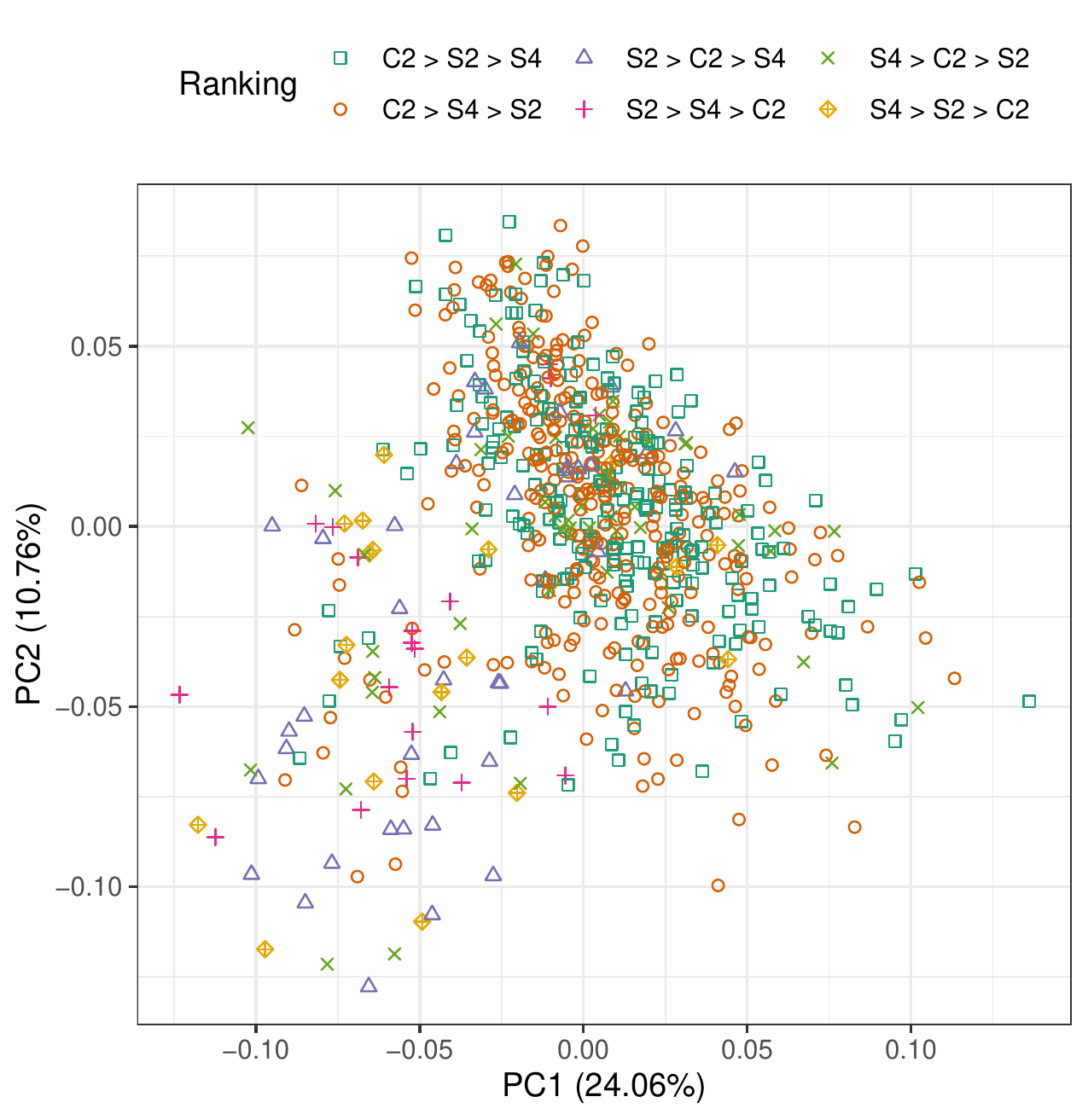}
\hfill
\includegraphics[width=0.32\linewidth]{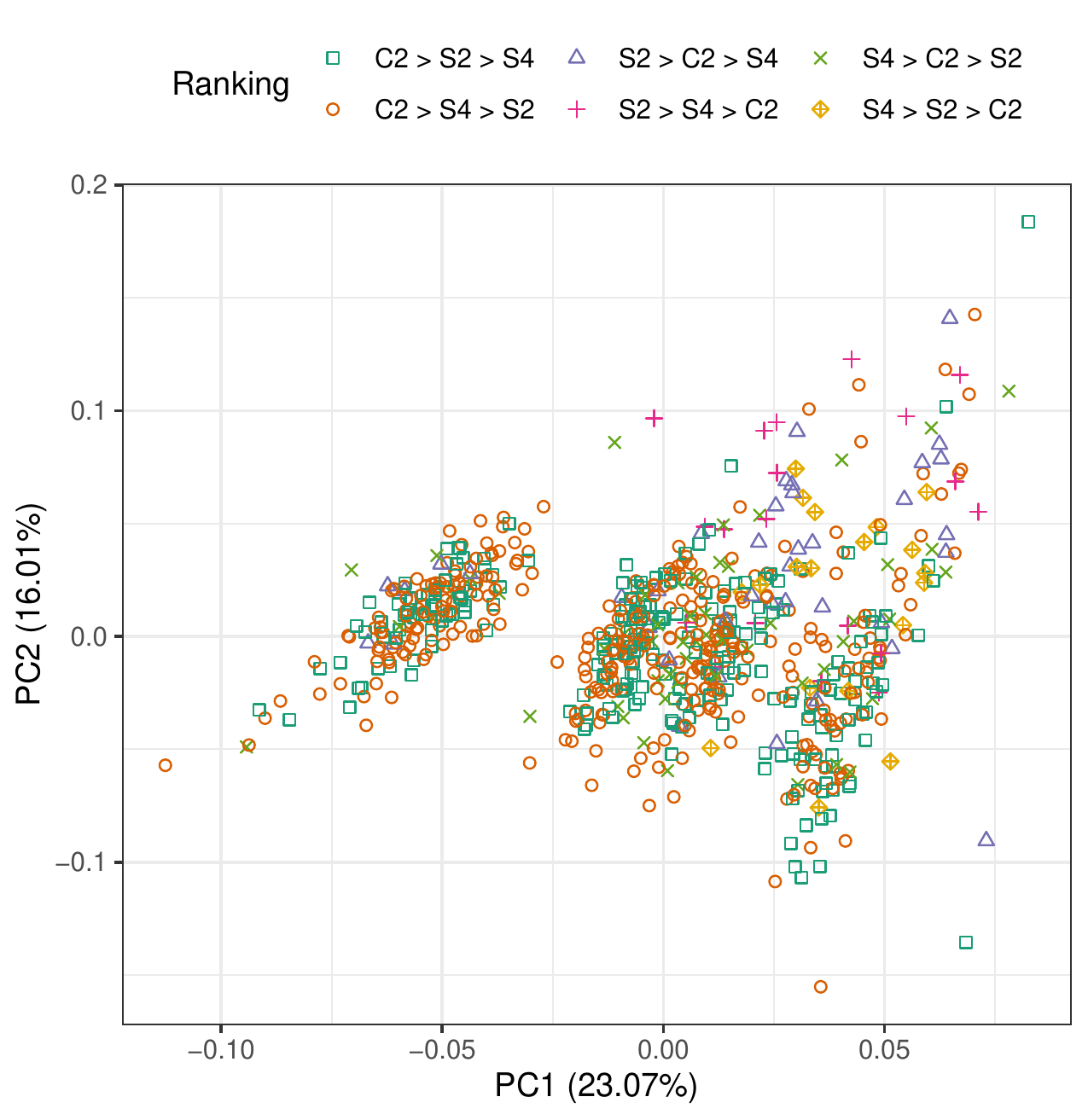}
\vskip-10pt
\caption{First two principal components results of a principal component analysis on both TSP and KP features (left), only TSP features (center) and only KP features (right).}
\label{fig:pca_on_features}
\end{figure*}

\begin{figure*}
\centering
\includegraphics[width=\columnwidth]{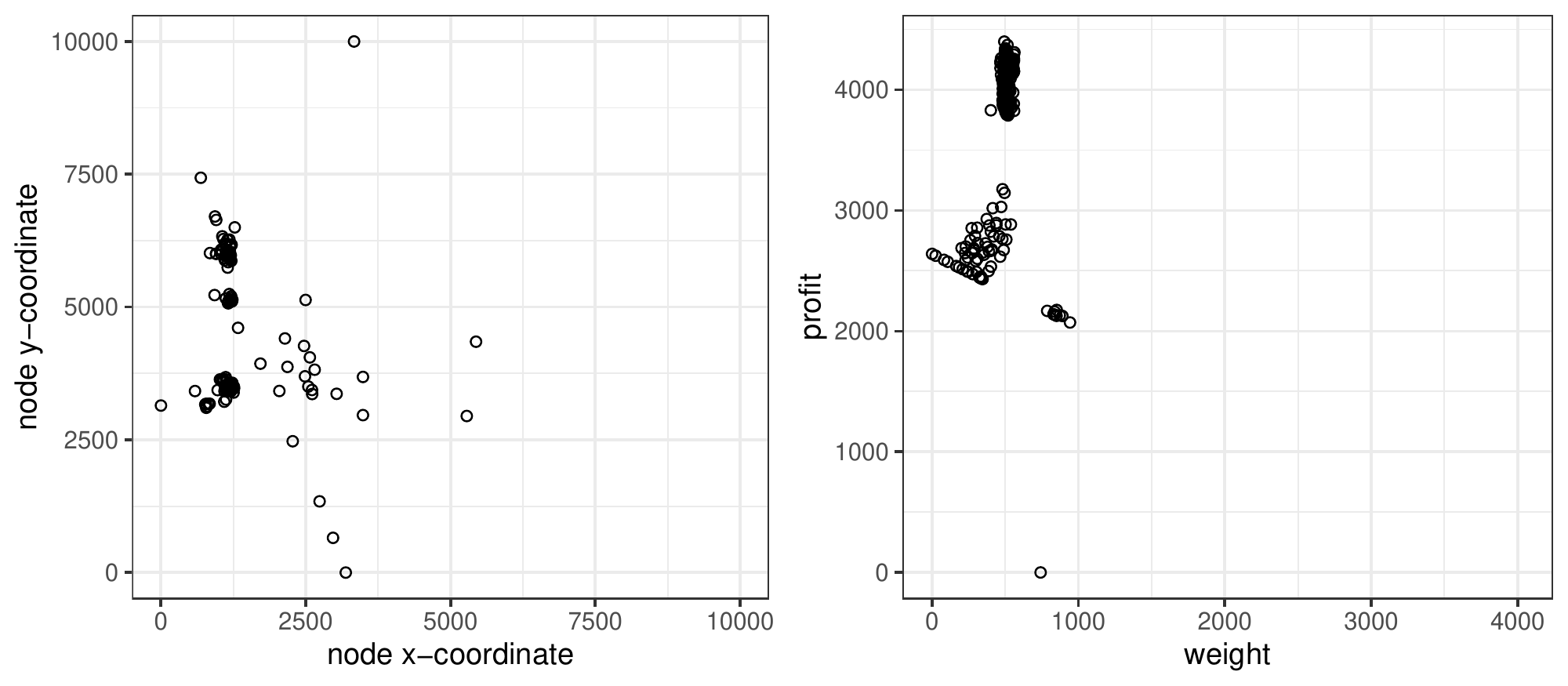}
\includegraphics[width=\columnwidth]{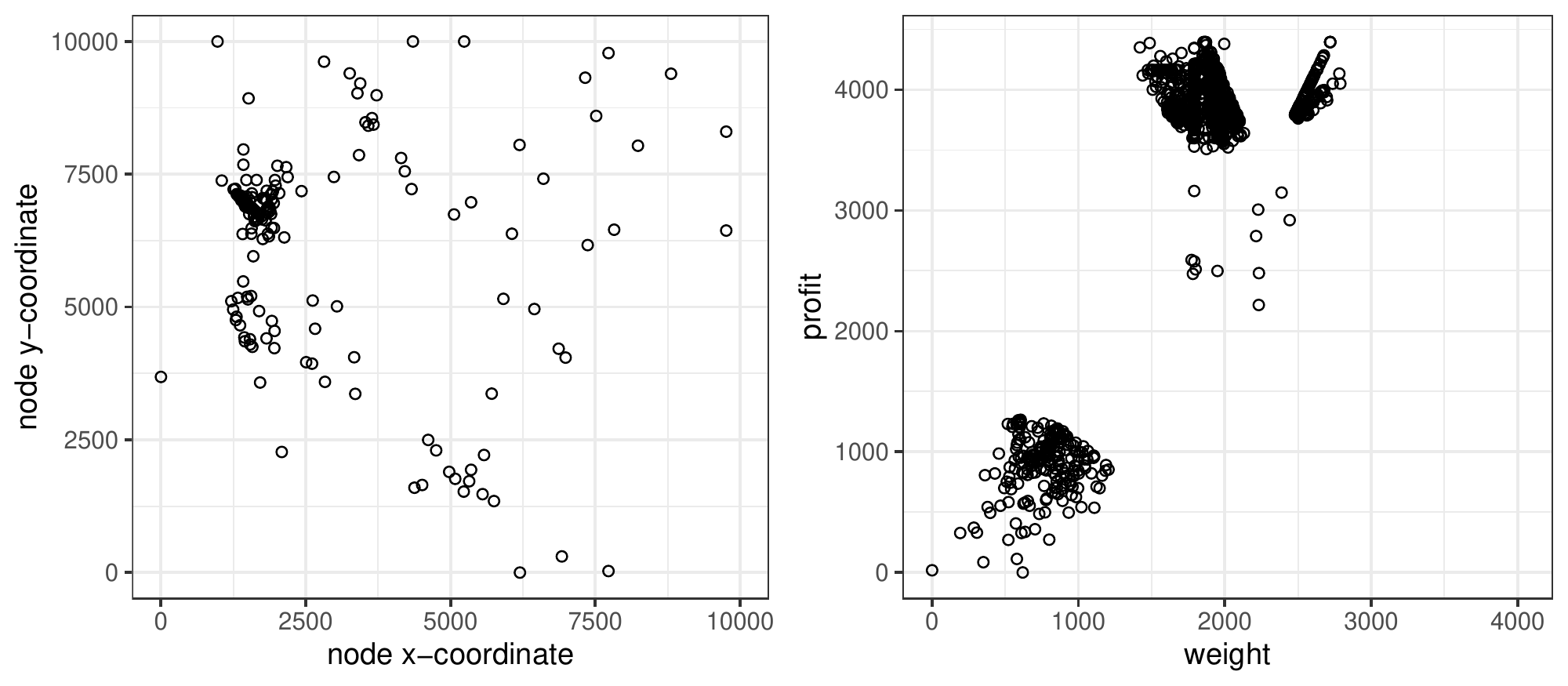}
\includegraphics[width=\columnwidth]{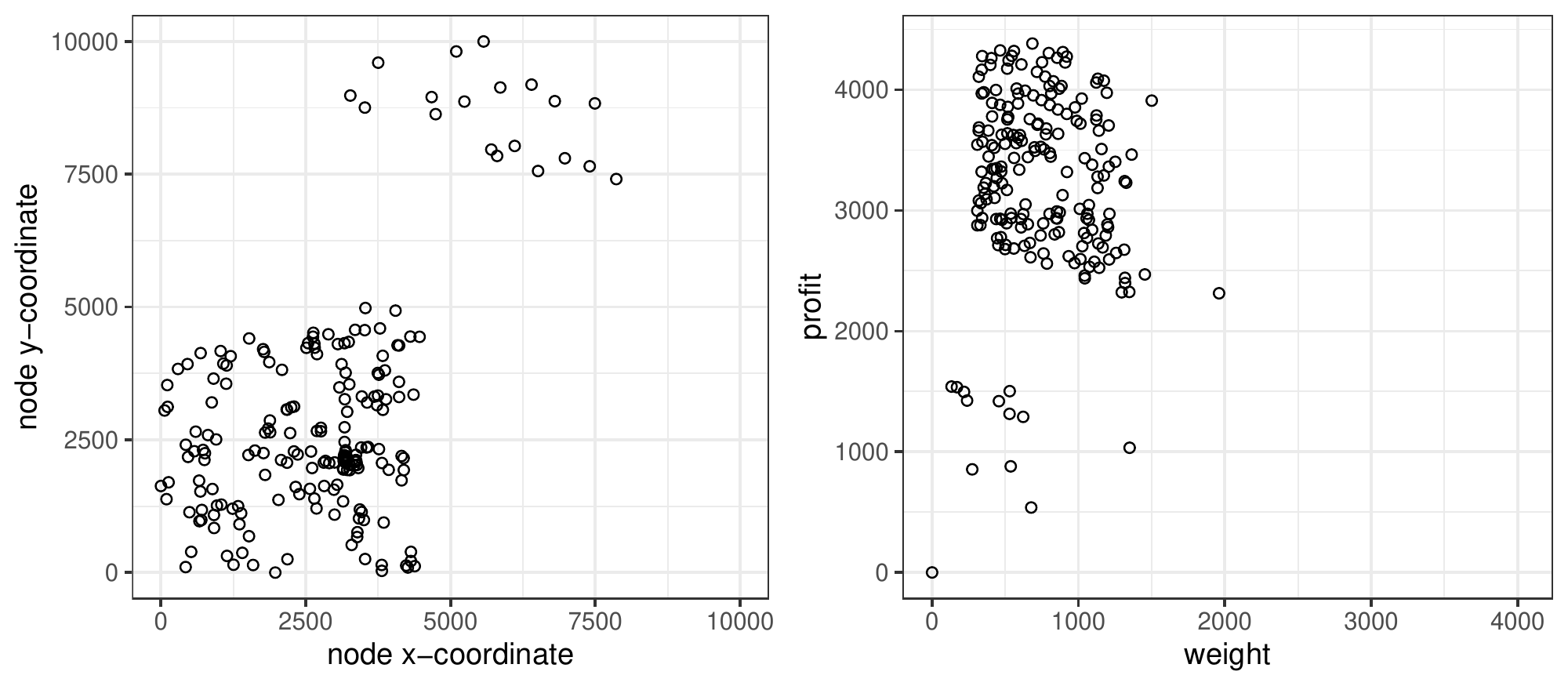}
\includegraphics[width=\columnwidth]{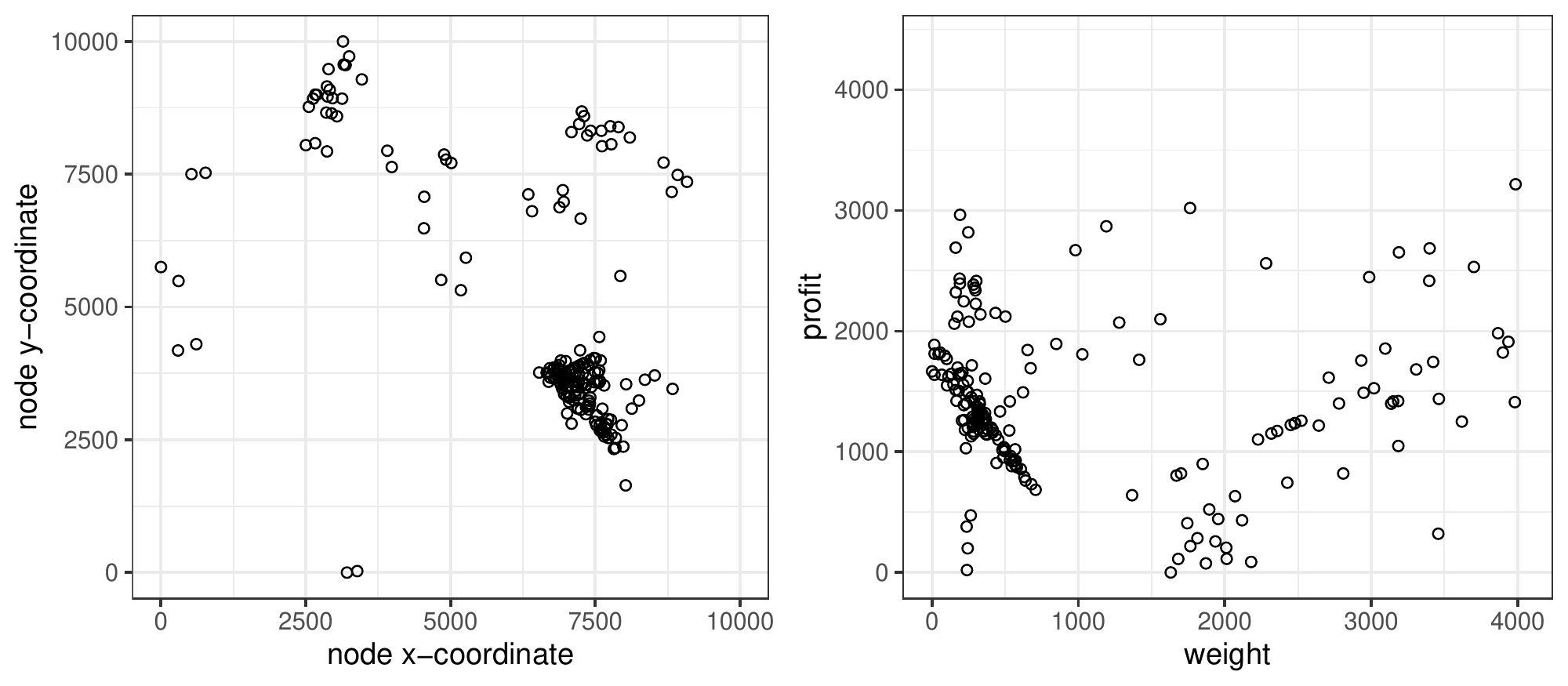}
\vskip-10pt
\caption{Evolved TTP instances where the desired ranking was achieved successfully for $S2 > S4 > C2$ (top-left), $S4 > C2 > S2$ (top right), $C2 > S2 > S4$ (bottom left) and $C2 > S4 > S2$ (bottom right).}
\label{fig:ttp_evolved_gallery}
\end{figure*}

To the best of our knowledge there is no work on instance characteristics for the TTP besides a brief investigation in the context of algorithm selection~\cite{Wagner2017JoH}. Therefore, we treat the TSP and KP components separately. For each instance we calculate a set of features for the TSP taken from the literature. Features involve summary statistics of the edge weights of a minimum spanning tree (e.g., the depth), distance-based features calculated on basis of the distance matrix (mean, median, standard deviation etc.) or properties of the $k$-Nearest-Neighbor Graph~(NNG) like the number of weak/strong components it is composed of. In particular the latter $k$-NNG based feature-set proved useful in algorithm selection approaches for the TSP~\cite{PM2014TSPfeatures}. The KP-related weight/profit combination is treated as another TSP and we calculate the same features for the knapsack component of the TSP. This approach results in a set of more than $400$ features in total given. We apply principal component analysis~(PCA) to the features and project the instances into the 2-dimensional space spanned by the first two components.\footnote{Prior to PCA, in order to obtain tidy numerical input data for the method, we removed constant features and rare cases where the calculation of certain features yielded infinity. These artifacts are due to the evolving process not avoiding duplicate node coordinates.} Figure~\ref{fig:pca_on_features} shows the 2-dimensional embeddings if PCA was applied with the union of TSP and KP features, TSP features only and KP-features only. The points are colored by their \emph{actual ranking} based on the final evaluation. Notably, in line with observations made at the beginning of the experimental evaluation, the majority of points represents instances that are easiest for C2 (cf.~Figure~\ref{fig:barplot_actual_ranking}). The total variance in the data explained by the first two principal components (PCs) is not tremendous, but neither is it low. We have $\approx 27.5\%$ in case we use both feature sets, $\approx 34.7$ for TSP-only features and $\approx 39\%$ for KP-only features. In the first and last plot we can identify four (partly overlapping) clusters which can be attributed to the four values $\{1, 3, 5, 10\}$ used for the number of items per node ($\mathit{IPN}$). This makes sense because many of the features are no normalized and thus affected by the number of observations (note that the bounding box for the items is $[0,4\,040] \times [0,4\,400]$ regardless of the choice for $\mathit{IPN}$). Taking a close look at the plots we can see that instances which are easiest for S2 are located in another area of the plot (violet triangles and orange dots) quite well separated from the large clusters where most C2-dominated instances lie.

For example, in the left-most embedding the majority of these instances have a PC1-value below $-0.025$. Making sense of the so-called loadings of the PCA is difficult. There are many features that influence the first and second principal component and it is out of scope of this paper to dive deeper into machine learning models. However, it opens a path for deeper instance-feature analysis and paves the way for a better understanding of TTP-instance hardness. We stress that these observations are particularly nice given the discouraging and unpromising results discussed earlier in this section.

Finally, Figure~\ref{fig:ttp_evolved_gallery} gives a visual impression of four representative evolved instances. The instances are very diverse owing the disruptive nature of the adopted mutation operators. The overall impression is that S2 works best on instances with rather densely clustered node weights and items, i.e., where there exist subsets of items with similar weights and profits. In contrast, C2 copes better with a wider weight/profit spread.


\section{Conclusion and take-away messages}
\label{sec:conclusion}

We have studied the task of generating a set of benchmark instances for combinatorial optimization problems by means of evolutionary algorithms. Such a set can aid researchers to get a better understanding of the problem and develop better algorithms. Ideally, such a set is diverse with respect to (1) complementary algorithm performance on a set of algorithms and (2) structural properties of the instances. We aimed for the first goal and targeted the problem of evolving instances where the performance on at least three algorithms follows a given ranking. To this end we proposed fitness functions suited to guide the evolutionary search process in a way that it is balanced with respect to the ranking of solver performance. As part of a proof-of-concept study we adopted our approach to evolve a benchmark set for the Traveling Thief Problem~(TTP) and three TTP-heuristics.

The results clearly show that the effectiveness of our approach strongly depends on the algorithm portfolio. As a take-away message we want to make the reader aware of the following potential pitfalls in this branch of research:
\begin{enumerate}
    \item Unsurprisingly, it is easy to evolve instances that are most reliably solved by the \emph{dominating} algorithm in the portfolio, but the reverse can be difficult, if not even impossible. Hence, the proposed portfolio necessarily needs to be composed of solvers with strong complementary behavior.
    \item The proposed fitness functions can be fooled by certain statistical artifacts of the solver performance. In this work, most notably, severe bi-modality of the performance distribution of single solvers led to a misdirection of the evolutionary search even though the robust median value was used to aggregate over multiple runs.
\end{enumerate}

In future work, we will apply our approach to other optimization problems, and we will improve the algorithm used for the evolution: while a population-based approach will enable us to evolve diverse instances in parallel, we will need to investigate how to evenly distribute them in the space of $N!$ target permutations. For this, established distance measures (see e.g.~\cite{Cicirello2018}) might prove useful.

Moreover, the formulated take-away messages suggest interesting avenues for further investigations, such as: how do task-specific mutation operator have to be constructed that are better suited to generate instances that are indeed hard for the dominating algorithm; how can alternative summary statistics (e.g., $p$-quantiles) assist to aggregate solver performance?

\vspace{2mm}\textbf{Acknowledgements.}
Jakob Bossek acknowledges support by the \href{https://www.ercis.org}{\textit{European Research Center for Information Systems (ERCIS)}}. Markus Wagner acknowledges support by the \textit{Australian Research Council projects} DP200102364 and DP210102670.

\clearpage
\bibliographystyle{ACM-Reference-Format}
\bibliography{arxiv}

\end{document}